\documentclass[11pt]{article}
\usepackage[letterpaper,margin=0.94in]{geometry}
\usepackage[T1]{fontenc}
\usepackage[utf8]{inputenc}
\usepackage{lmodern}
\usepackage{microtype}
\usepackage{amsmath,amssymb,mathtools,amsthm}
\usepackage{natbib}
\setcitestyle{authoryear,round,citesep={;}}
\usepackage{booktabs,tabularx,array,multirow}
\usepackage{graphicx}
\usepackage[section]{placeins}
\usepackage{float}
\usepackage{enumitem}
\usepackage{xcolor}
\usepackage{hyperref}
\usepackage[nameinlink]{cleveref}
\hypersetup{colorlinks=true,citecolor=blue!45!black,linkcolor=blue!45!black,urlcolor=blue!45!black,
  pdftitle={When Exact Token Importance Is Not Enough: Limits of Top-K Selection in LLM Attention},
  pdfauthor={Mohammad Mirzanejad}}
\newtheorem{proposition}{Proposition}
\newtheorem{theorem}{Theorem}

\newcommand{\KL}{D_{\mathrm{KL}}}
\newcommand{\Swap}{\textsc{SwapCert}}
\newcommand{\TopK}{\operatorname{Top}\!\mbox{-}K}
\newcommand{\E}{\mathbb{E}}
\newcommand{\cF}{\mathcal{F}}
\newcommand{\cN}{\mathcal{N}}

\setlist[itemize]{nosep,leftmargin=1.3em}
\title{When Exact Token Importance Is Not Enough: \\ Limits of Top-K Selection in LLM Attention}
\author{
	Irmaan Mirzanejad\thanks{The author's legal given name
		is Mohammad. He publishes under his preferred given
		name, Irmaan.}\\
	GREYC, CNRS, Université de Caen Normandie
}
\date{}
\begin{document}
\maketitle
\begin{abstract}
Long-context language models often retain only a limited set of past token positions for an attention query. A common strategy scores positions individually and keeps the highest-ranked $K$. Yet the attention weights are renormalized over the chosen set: two individually important positions need not form the best pair. We ask whether independent ranking suffices even when each importance score measures the \emph{exact downstream change in next-token predictions} caused by deleting that position alone. It does not. A four-position construction makes the ranked pair incur $133.2\times$ the Kullback--Leibler (KL) divergence of the optimal pair, and a parametric family proves that this multiplicative gap is unbounded. An exact exchange identity and preference reversals for every $K\ge2$ explain the interaction. In exhaustive frozen-model audits of SmolLM2, Gemma, and Qwen, exact-singleton Top-4 leaves at least $10\%$ relative improvement within a fully enumerated 24-position candidate pool in 55/60, 49/60, and 60/60 cases, respectively. \Swap{}, a same-budget one-swap audit, returns a directly checkable improving support and detects 157/164 such cases. A KL--margin certificate relates support fidelity to next-token stability; larger-budget, multi-layer cached-decoding experiments measure predictive fidelity at $K=128$ and $256$. These results establish \emph{joint support quality} as an essential object of analysis for sparse attention and motivate downstream-aware support selection under inference constraints.
\end{abstract}

\section{Introduction}\label{sec:introduction}
Consider a large language model (LLM) answering a question about a long document. To reduce attention cost at a decoding step, it may attend to only $K$ earlier token positions. A natural shortcut ranks the positions one by one and retains the $K$ highest-scoring ones. The complication is that attention is a \emph{weighted combination}: after positions are discarded, softmax recomputes the relative weights of all survivors. Two positions that appear highly influential when tested separately can therefore pull the retained representation away from the model's full-attention representation. A lower-ranked position may instead complete a much better set.

\paragraph{A four-position example.}
The effect already occurs in a small, exactly solvable attention head. Its full-attention output is $2.15$. Measuring each of four positions by the exact change in downstream Kullback--Leibler (KL) divergence when that position alone is deleted ranks positions $4$ and $1$ highest. But keeping only $\{1,4\}$ produces an output of $3.071$, while the lower-ranked pair $\{3,4\}$ produces $2.214$, much closer to the full-attention output. After the same frozen binary next-token predictor, the first pair has \emph{$133.2$ times the downstream KL} of the second (Section~\ref{sec:counterexample}). The example illustrates the central distinction: \emph{The surprising implication is that even perfect knowledge
	of individual token importance can lead to a substantially
	inferior retained set. The failure arises from the selection
	rule itself, rather than from errors in estimating importance.
	As we subsequently prove, this discrepancy can be arbitrarily
	large in relative terms.}.

\paragraph{Why this matters for long-context inference.}
As context grows, attention requires increasing key--value (KV) cache storage, memory traffic, or both. Sparse-attention and KV-retention systems limit the active history through eviction, retrieval, query-conditioned masks, and related mechanisms \citep{Zhang2023H2O,Xiao2024Streaming,Ribar2024SparQ,Li2024SnapKV,Tang2024Quest}. Many such systems construct a support by assigning a scalar score to each position and selecting the Top-$K$. Improving a scorer is valuable, but the resulting support is jointly renormalized. The central problem is therefore: \emph{which set of exactly $K$ positions best preserves the downstream next-token distribution?}

\paragraph{An exact-score test of the ranking abstraction.}
We isolate this problem from errors in importance estimation. Imagine an oracle that measures the actual change in the frozen language model's next-token distribution whenever one position is deleted from the full attention support. These are exact, objective-matched singleton-deletion scores. We ask whether ranking them is sufficient for fixed-budget support selection. The four-position counterexample shows a large finite loss gap; a continuous family strengthens this to \emph{unbounded multiplicative suboptimality}. In addition, an exact exchange identity demonstrates that the outcome of replacing one retained position depends on the current set's total weight and weighted output. We prove that the preference between the same two candidates can reverse when their accompanying retained positions change, at every budget $K\ge2$.

\paragraph{From the structural result to a measurable diagnostic.}
For a fixed support budget, we call the reduction in downstream loss achievable by choosing a better support \emph{headroom}. We measure it against exact global optima inside a \emph{candidate universe}: a prespecified pool of $M=24$ eligible positions from which the retained $K$ are selected. Complete enumeration is feasible in this controlled setting. Across frozen full-softmax attention heads from three LLM families, exact downstream-singleton Top-4 frequently leaves at least $10\%$ relative headroom. We measure its absolute KL scale and the conditioned exchange preferences underlying the gap. \Swap{} then examines the one-exchange neighborhood of any incumbent support. An issued witness identifies an incoming position, an outgoing position, and a verifiable decrease in the same downstream loss. We calibrate its empirical detection rate against the fully enumerated optimum. The contribution of this diagnostic is its transparent measurement of remaining support headroom; the mathematical novelty lies in the objective-matched ranking/support separation.

\paragraph{Connecting support quality to model behavior.}
We also examine the predictive consequences of changing a support. A KL--margin condition guarantees preservation of the dense model's top next-token prediction whenever the downstream divergence is sufficiently small. A standard autoregressive KL chain rule extends the interpretation to distributions over generated sequences. At larger active budgets $K\in\{128,256\}$, we measure multi-layer cached-decoding fidelity using a practical geometric score, attention-based selection, and position-based controls. We report dense-reference KL, reference-token negative log-likelihood (NLL), and observed generated continuations. These complementary measures distinguish distributional fidelity, discrete predictions, and task outcomes under clearly specified interventions.

\paragraph{Contributions.}
\begin{enumerate}[leftmargin=1.5em,itemsep=2pt]
	
	\item \textbf{Fundamental limits of exact token ranking.}
	We establish that even exact singleton importance,
	measured under the target downstream prediction objective,
	can yield arbitrarily inferior fixed-budget supports.
	A constructive counterexample and an unbounded
	approximation-gap theorem establish this limitation
	already at $K=2$ (Section~\ref{sec:theory}).
	
	\item \textbf{Mathematical characterization of support dependence.}
	We derive an exact support-exchange identity and prove
	that preferences between candidate positions can reverse
	as a function of their retained context for every
	$K\geq2$. These results characterize the structural
	mechanism underlying the separation between individual
	importance and joint support quality
	(Section~\ref{sec:exchange}).
	
	\item \textbf{Exact empirical characterization and certified
		improvement.}
	Through exhaustive support evaluation in three LLM families,
	we quantify the prevalence and magnitude of the ranking--support
	gap, examine its underlying mechanism, and validate
	\Swap{} as a constructive certificate of available
	same-budget improvement. We further connect support quality
	to predictive decisions through KL--margin certification
	and larger-budget, multi-layer decoding experiments
	(Sections~\ref{sec:swapcert}--\ref{sec:results}).
	
\end{enumerate}

Together, these contributions establish a theoretical
and empirical foundation for \emph{downstream-aware support
	selection under computational, memory, and temporal
	constraints}, connecting the fundamental limitations of
independent ranking with the broader problem of joint
support optimization in LLMs.

\section{Related Work}\label{sec:related}
\paragraph{Ranking, retrieval, and KV budgets.}
Heavy-Hitter Oracle (H2O) combines heavy-hitter and recent-token retention \citep{Zhang2023H2O}; StreamingLLM maintains attention sinks and a recent window \citep{Xiao2024Streaming}; SparQ and Quest selectively retrieve KV history \citep{Ribar2024SparQ,Tang2024Quest}; and SnapKV uses prompt-derived per-head selection \citep{Li2024SnapKV}. Ada-KV optimizes the allocation of retention budgets across attention heads \citep{Feng2025AdaKV}. These systems operate at different granularities and cost models. The common evaluation question is the quality of the support actually activated by a particular query.

\paragraph{Value-aware scores and conditional guarantees.}
CriticalKV analyzes output perturbations beyond raw attention mass \citep{Feng2026Critical}. AnDPro motivates token scoring through value-space geometry \citep{Geng2025AnDPro}. DropKV studies residual-output perturbation and derives a constant-factor selection guarantee under an explicit cone condition \citep{Zhang2026DropKV}. These positive results illustrate when economical scores can be justified by structure. We instead test an assumption-free implication for the \emph{same downstream KL objective} that evaluates the final set: even granting exact singleton-deletion KL, can independent Top-$K$ optimize the jointly renormalized support? Our unbounded-ratio result establishes why useful approximation guarantees require additional structure.

\paragraph{Set-aware selection and exact subset reconstruction.}
BumbleBee explicitly formulates streaming KV summarization through submodular subset selection \citep{Kumari2024Bumble}; AnDPro also begins from a retained-set objective. Trust the Mass examines forced renormalization, exact restricted subset optimization, and head-output reconstruction \citep{Shi2026Mass}. Our study is complementary in both \emph{objective} and \emph{oracle}: the evaluated loss is downstream next-token KL after the frozen Transformer remainder, and the ranker is supplied exact downstream singleton effects. The exhaustive candidate-universe optimum calibrates how much that specific ranking can leave on the table. Head-output reconstruction and downstream-KL support losses can exhibit different relative gaps; the absolute scale and outcome metrics must be considered alongside relative headroom.

\begin{table}[H]
\centering\small
\caption{\textbf{Related methodological perspectives.} The fidelity object and type of guarantee distinguish support-level diagnosis from deployed KV selection.}
\label{tab:related-comparison}
\setlength{\tabcolsep}{5pt}
\begin{tabularx}{\linewidth}{@{}>{\raggedright\arraybackslash}p{.24\linewidth}>{\raggedright\arraybackslash}p{.31\linewidth}>{\raggedright\arraybackslash}X@{}}\toprule
Approach & Object of evaluation & Mathematical or computational role \\\midrule
Value-aware selectors & Head-output perturbation & Efficient approximations of token importance \\
DropKV & Renormalized head-output residual & Conditional approximation guarantee \\
Set-aware methods & Subset utility and reconstruction & Support-dependent optimization \\
Exact singleton oracle (this work) & Downstream KL of each deletion & Objective-matched ranking sufficiency test \\
SwapCert (this work) & Downstream KL of exchanged support & Checkable one-sided headroom witness \\\bottomrule
\end{tabularx}
\end{table}

\section{Fixed-Budget Support Selection and Theory}\label{sec:theory}
\subsection{Softmax supports and an objective-matched singleton oracle}
Let $D$ denote the set of valid past positions in a frozen softmax-attention head \citep{Vaswani2017Attention}, $s_i$ the unnormalized attention score (logit) of position $i$, and $v_i$ its value vector. With every valid position retained, the dense weights and head output are
\begin{equation}\label{eq:dense}
a_i=\frac{e^{s_i}}{\sum_{j\in D}e^{s_j}},\qquad h_D=\sum_{i\in D}a_i v_i.
\end{equation}
For a retained support $S\subseteq D$, sparse-softmax renormalization yields
\begin{equation}\label{eq:softmaxsupport}
A_S=\sum_{i\in S}a_i,\qquad h_S=\frac{\sum_{i\in S}a_i v_i}{A_S}.
\end{equation}
Here $A_S$ is the original dense attention mass retained by $S$, and $h_S$ is the resulting renormalized head output.
Let $p_D$ be the dense model's next-token distribution and $p_S$ the distribution after replacing the target head output with $h_S$ and propagating through the unchanged downstream model. The joint support loss is
\begin{equation}\label{eq:loss}
L(S)=\KL(p_D\Vert p_S).
\end{equation}
For a prespecified candidate universe $U\subseteq D$ of $M=|U|$ eligible positions and a support budget $1\le K<M$, define $\cF_K(U)=\{S\subseteq U:|S|=K\}$ and $S_U^\star\in\arg\min_{S\in\cF_K(U)}L(S)$. To eliminate importance-estimation error, give each candidate its \emph{exact downstream singleton-deletion loss}
\begin{equation}\label{eq:singleton}
d_j=\KL\bigl(p_D\Vert p_{D\setminus\{j\}}\bigr),\quad j\in U.
\end{equation}
Write $S_{\mathrm{rank}}=\TopK(\{d_j\}_{j\in U})$ for the $K$ highest-scoring candidates (with a fixed tie-breaking rule). The central question is whether $S_{\mathrm{rank}}$ always minimizes \eqref{eq:loss} over $\cF_K(U)$.
\paragraph{Three quantities with different meanings.}
The \emph{singleton score} $d_j$ evaluates one deletion from the dense context. The \emph{support loss} $L(S)$ evaluates a jointly renormalized retained set. The \emph{headroom} of an incumbent support measures how much lower that same support loss can become at fixed cardinality. The distinction concerns their mathematical relationship, independently of any particular ranking heuristic. For an open-ended prompt, $p_D$ is simply the frozen model's next-token distribution at the observed prefix; its KL against $p_S$ is computable without a reference answer (Eq.~\ref{eq:logscore}).

\subsection{An exact objective-matched counterexample}\label{sec:counterexample}
Consider four dense weights $a=(0.2,0.1,0.2,0.5)$, scalar values $v=(4,-2,1,2.7)$, and downstream binary logits $z(h)=(1.8h,0)$. The dense head output is $h_D=2.15$. Direct evaluation of all four singleton-deletion KL values gives
\[
d_4=0.01374412>d_1=0.00918668>d_2=0.00536907>d_3=0.00228497.
\]
Thus exact-singleton Top-2 selects $\{1,4\}$. Exact enumeration of all six size-2 supports gives $S_U^\star=\{3,4\}$, with
\begin{equation}\label{eq:counterexample}
L(\{1,4\})=0.01720822267,\qquad L(\{3,4\})=0.000129168300,
\end{equation}
a $133.2$-fold loss ratio.
\begin{proposition}[Exact singleton ranking is insufficient]\label{prop:impossibility}
Even exact, objective-matched downstream singleton-deletion KL scores do not generally identify a jointly optimal fixed-cardinality support under softmax renormalization.
\end{proposition}
The values above establish the proposition by construction; all singleton and pair evaluations appear in Appendix~\ref{app:counterexample}. Figure~\ref{fig:separation} makes the central distinction visible: the singleton ordering is exact, yet the two selected tokens form a poor pair after renormalization.
\begin{figure}[!htbp]
\centering
\includegraphics[width=.97\linewidth]{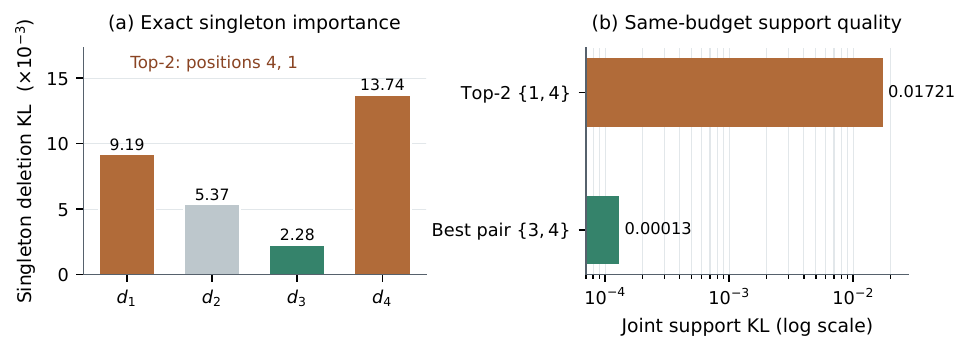}
\caption{\textbf{Exact marginal ranking versus joint support quality.} Left: objective-matched downstream singleton-deletion KL ($d_j$). Right: downstream KL of the two size-2 supports. Exact singleton Top-2 selects $\{1,4\}$, while the optimal pair is $\{3,4\}$; the loss ratio is $133.2\times$. Both supports use the same cardinality and the same dense reference distribution.}
\label{fig:separation}
\end{figure}

\subsection{An unbounded approximation gap}\label{sec:unbounded}
The finite $133.2\times$ example has strictly positive optimum and captures an experimentally checkable case. The structural limitation is stronger: the ranked support can have arbitrarily worse downstream KL than the optimal support under the same budget.
\begin{proposition}[No uniform multiplicative approximation for exact-singleton Top-2]\label{prop:unbounded}
For every $R>0$, there exists a four-position softmax head with a binary frozen downstream predictor such that the exact downstream singleton-deletion scores have a strict Top-2 ordering, the best size-two support has positive loss, and
\[
\frac{L(\TopK(\{d_j\},2))}{\min_{|S|=2}L(S)}>R.
\]
Consequently, exact objective-matched singleton Top-$K$ has no instance-uniform multiplicative approximation factor over this class, already at $K=2$.
\end{proposition}
\begin{proof}[Proof sketch]
Keep the rational weights $a=(1/5,1/10,1/5,1/2)$ and binary logits $(1.8h,0)$, and set $v=(4,-2,1/4+\delta,27/10)$. At $\delta=0$, the dense output and the output of support $\{3,4\}$ both equal $2$, while exact singleton scores strictly order the candidates as $d_4>d_1>d_2>d_3$. Hence the ranked support is $\{1,4\}$ and has strictly positive KL. For sufficiently small $\delta>0$, continuity preserves the strict singleton ordering. Meanwhile $h_{\{3,4\}}-h_D=3\delta/35$, so the KL of this feasible support tends to zero quadratically, whereas the KL of ranked support $\{1,4\}$ tends to a positive constant. All six size-two supports have positive KL for sufficiently small $\delta>0$. Thus the ratio exceeds any $R$. Appendix~\ref{app:unbounded-proof} gives the exact inequalities and an independent verifier.
\end{proof}
\begin{figure}[tb]
\centering
\includegraphics[width=.80\linewidth]{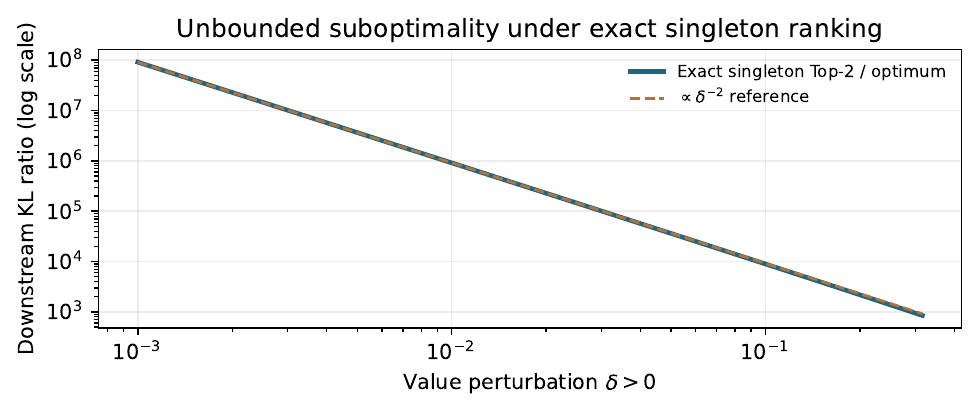}
\caption{\textbf{Unbounded loss ratio with a fixed exact singleton ordering.} The four-token perturbation family preserves the Top-2 support $\{1,4\}$ while the jointly optimal pair $\{3,4\}$ approaches the dense downstream prediction. The ratio diverges as $\delta\downarrow0$. The curve illustrates the exact construction established analytically in Proposition~\ref{prop:unbounded}.}
\label{fig:unbounded}
\end{figure}

This establishes a precise boundary between \emph{measuring singleton relevance perfectly} and \emph{approximating support quality uniformly}. Structural assumptions such as those underlying positive approximation guarantees in related work are therefore substantively meaningful.

\subsection{Support-conditioned exchange value}\label{sec:exchange}
For $S'=S\setminus\{i\}\cup\{j\}$, with $i\in S$ and $j\notin S$, the restricted outputs satisfy
\begin{align}\label{eq:exchange1}
h_{S'}&=\frac{A_S h_S-a_iv_i+a_jv_j}{A_S-a_i+a_j},\\[-0.3em]
\label{eq:exchange2}
h_{S'}-h_S&=\frac{a_j(v_j-h_S)-a_i(v_i-h_S)}{A_S-a_i+a_j}.
\end{align}
The exact displacement depends jointly on retained mass $A_S$, current output $h_S$, and both exchanged positions. For fixed $K$, the resulting downstream preference is conditioned on the accompanying support.
\begin{proposition}[Preference reversal for every $K\ge2$]\label{prop:reversal}
For every fixed support budget $K\ge2$, the same pair of candidate positions can reverse downstream-KL preference solely because the other retained positions change. A context-independent scalar ranking cannot encode every fixed-budget exchange preference.
\end{proposition}
\begin{proof}
Use uniform attention on $K+2$ values consisting of $K-2$ zeros and the four values $-2,+2,+1,-1$. Let the shared $K-1$-element context contain the $K-2$ zeros and either $-2$ or $+2$. Under a symmetric binary downstream map with dense output $0$, KL is strictly increasing in the absolute retained output. In the negative context, appending $+1$ gives $-1/K$, while appending $-1$ gives $-3/K$; $+1$ is preferred. In the positive context, the corresponding outputs are $3/K$ and $1/K$; $-1$ is preferred. The two candidate preferences therefore reverse for every $K\ge2$.
\end{proof}

\paragraph{A certified exchange in the four-token construction.}
The counterexample makes the exchange geometry directly observable. Starting from exact-singleton Top-2, $S=\{1,4\}$, replacing $i=1$ by $j=3$ preserves the support budget and changes the restricted output from
$h_{\{1,4\}}=3.071429$ to $h_{\{3,4\}}=2.214286$.
The corresponding downstream loss decreases by $0.01707905$ (a $99.25\%$ relative reduction). Thus one verified same-budget exchange connects the objective-matched ranking failure to the constructive witness returned by \Swap{}; the exact four-token optimum is attained in this single step.

\subsection{Predictive fidelity and next-token stability}\label{sec:kl-margin}
For any frozen prefix, the downstream KL is the expected log-score regret under the dense predictive distribution:
\begin{equation}\label{eq:logscore}
\KL(p_D\Vert p_S)=\E_{Y\sim p_D}[-\log p_S(Y)]-\E_{Y\sim p_D}[-\log p_D(Y)].
\end{equation}
The reference distribution is defined directly by the frozen model at that prefix, including open-ended questions. This identifies a label-free objective for \emph{predictive fidelity}; reference-answer NLL offers a complementary, task-related measurement.
\begin{proposition}[KL--margin top-token guarantee]\label{prop:margin}
Let $y^*=\arg\max_y p_D(y)$ be unique with margin $m=p_D(y^*)-\max_{y\ne y^*}p_D(y)>0$. If $\KL(p_D\Vert p_S)<m^2/2$, then $\arg\max_y p_S(y)=y^*$.
\end{proposition}
\begin{proof}
Pinsker's inequality gives $\|p_D-p_S\|_1\le\sqrt{2\KL(p_D\Vert p_S)}<m$. For $y\ne y^*$, $p_S(y^*)-p_S(y)\ge m-|p_S(y^*)-p_D(y^*)|-|p_S(y)-p_D(y)|\ge m-\|p_S-p_D\|_1>0$.
\end{proof}

\paragraph{From one step to a generated sequence.}
Let $P_D^{(T)}$ and $P_\pi^{(T)}$ denote dense and support-policy distributions over $T$ generated tokens, with conditional predictions defined at the same prefixes. The standard autoregressive KL chain rule gives
\begin{equation}\label{eq:trajectory-kl}
\KL(P_D^{(T)}\Vert P_\pi^{(T)})=\sum_{t=1}^{T}\E_{Y_{<t}\sim P_D}\!\left[\KL\bigl(p_D(\cdot\mid Y_{<t})\Vert p_\pi(\cdot\mid Y_{<t})\bigr)\right].
\end{equation}
For any measurable event $A$ on complete continuations---including an answer satisfying a specified correctness rule---Pinsker's inequality implies
\begin{equation}\label{eq:event-control}
\big|P_D^{(T)}(A)-P_\pi^{(T)}(A)\big|\le\sqrt{\tfrac12\KL(P_D^{(T)}\Vert P_\pi^{(T)})}.
\end{equation}
These classical identities clarify how fixed-prefix fidelity can control sequence-level distribution changes when evaluated along the dense-prefix distribution. They also distinguish \emph{closeness to the dense model} from \emph{improving benchmark accuracy}, which requires direct outcome evaluation. A proof and the required conditional-distribution assumptions are given in Appendix~\ref{app:sequence-bridge}.

\section{SwapCert: Constructive Same-Budget Headroom Auditing}\label{sec:swapcert}
Define $\cN_1(S)=\{S\setminus\{i\}\cup\{j\}:i\in S,j\in U\setminus S\}$. For $L(S)>0$, the global candidate-universe headroom and best one-swap witness are
\begin{equation}\label{eq:headroom}
H_U(S)=1-\frac{\min_{T\in\cF_K(U)}L(T)}{L(S)},\qquad
W_U(S)=\max\!\left\{0,1-\frac{\min_{T\in\cN_1(S)}L(T)}{L(S)}\right\}.
\end{equation}
For $L(S)=0$, both are set to zero. At threshold $\varepsilon$, \Swap{} evaluates $K(M-K)$ neighbor supports, returns the best exchange and its evaluated KL, and fires if $W_U(S)\ge\varepsilon$.
\begin{theorem}[One-sided witness soundness]\label{thm:soundness}
For any feasible support and $\varepsilon\in(0,1)$, $W_U(S)\ge\varepsilon$ implies $H_U(S)\ge\varepsilon$.
\end{theorem}
\begin{proof}
Since $\cN_1(S)\subseteq\cF_K(U)$, the minimum of $L$ over $\cF_K(U)$ is no greater than its minimum over $\cN_1(S)$. Hence $H_U(S)\ge W_U(S)$.
\end{proof}
The mathematical result concerns the ranking/support boundary. \Swap{} operationalizes it as a verifiable diagnostic: its witness is the evaluated improvement itself, while its recall is an empirical property of the audited candidate universes. At $M=24,K=4$, a certificate inspects $K(M-K)=80$ equally sized candidate supports, compared with the $10{,}626$ supports evaluated by the exact offline oracle. Its reported improvement is directly checkable by reevaluating the original and exchanged supports under the same downstream KL.

\paragraph{Oracle-call complexity and diagnostic workflow.}
The exact singleton oracle in Eq.~\eqref{eq:singleton} evaluates $M$ one-deletion interventions to build an objective-matched ranking. Once an incumbent $K$-support is available, exhaustive one-exchange auditing evaluates $K(M-K)$ restricted supports; the full candidate-universe oracle requires $\binom{M}{K}$ evaluations. At the primary $M=24,K=4$ setting this is $80$ versus $10{,}626$ support evaluations, a factor of $132.8$ in candidate count. Each loss evaluation consists of a fixed-head intervention followed by downstream model propagation, so evaluation count and total model-forward cost are distinct accounting units. A positive witness reports the entering and leaving positions and two evaluated predictive distributions, which permits direct independent verification. These computational quantities explain how exhaustive calibration and local certification complement each other within a controlled frozen-model audit.

\section{Experimental Design}\label{sec:methods}
\paragraph{Primary exact audit.} We evaluate HuggingFaceTB/SmolLM2-1.7B-Instruct, google/gemma-2-2b, and Qwen/Qwen3.5-4B with 30 frozen examples per family from LongBench-E, a long-context evaluation benchmark (ten each from the 2WikiMultihopQA and HotpotQA question-answering tasks and the English PassageRetrieval-en task \citep{Bai2024LongBench}). We intervene on the first and last heads of the deepest ordinary full-softmax layer (zero-indexed layers 23, 25, and 31). For each head, $U$ is the $M=24$ largest valid dense-attention positions. The support contains the $K=4$ highest scores $d_j$ from \eqref{eq:singleton}. Every one of the $\binom{24}{4}=10{,}626$ feasible supports is evaluated, totaling 1,912,680 support evaluations over 180 primary prompt--head cells. The head-output displacement is injected into the frozen downstream model. The $\varepsilon=0.10$ headroom threshold was fixed prior to outcome evaluation. Prompt-cluster bootstrap confidence intervals (CIs) resample prompts and retain both heads from each prompt.
\paragraph{Supporting exact audits.} Five outcome-independent candidate-universe constructions test universe sensitivity; exact $K\in\{2,3\}$ panels and a descriptive $K=5$ panel test budget sensitivity. Additional approximately 2K- and 4K-context tests screen prespecified same-budget exchanges across more heads (Appendices~\ref{app:universes}--\ref{app:larger-screens}).
\paragraph{Larger-budget cached decode.}\label{sec:v3-protocol}
The new protocol uses 2,048-token inputs, $K\in\{128,256\}$, and all heads in the last three eligible full-softmax layers. Model prefill is dense; selected attention is sparsified during subsequent single-query cached decoding. At each active step we compare attention-weight Top-$K$ with a geometric singleton score
\begin{equation}\label{eq:geometric}
u_j=\|h_{D\setminus\{j\}}-h_D\|_2=\frac{a_j}{1-a_j}\|v_j-h_D\|_2.
\end{equation}
This is the exact \emph{head-output} singleton displacement magnitude; the primary impossibility and exhaustive experiments use the distinct \emph{downstream-KL} singleton score $d_j$. We evaluate dense-reference KL, reference-answer NLL, recency and sink--recency controls, and matched positive-exchange screens. Precision and backend parity follow model-specific validated protocols; details are in Appendix~\ref{app:v3-evidence}. The intervention changes the active per-query attention support over a resident KV cache; the full cache remains available to the evaluator. Thus the experiment isolates predictive consequences of support choice, while physical memory reduction, cache eviction and end-to-end serving latency define a separate systems evaluation.

\section{Results}\label{sec:results}
\subsection{Exact singleton ranking leaves substantial candidate-universe headroom}
\begin{figure}[!htbp]
\centering
\includegraphics[width=.98\linewidth]{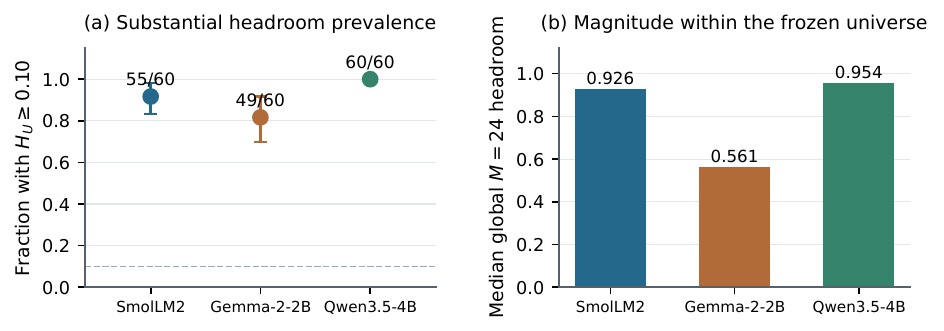}
\caption{\textbf{Primary exact support-selection audit at $M=24,K=4$.} Left: proportion of prompt--head cells whose exact downstream-singleton Top-4 support has at least 10\% relative headroom; whiskers resample the 30 prompt clusters, retaining both heads within each prompt. Right: median relative headroom against the exact best size-4 support in the frozen 24-position universe.}
\label{fig:primary}
\end{figure}

Figure~\ref{fig:primary} and Table~\ref{tab:primary} report the primary $K=4$ audit. Proposition~\ref{prop:unbounded} describes the worst-case theoretical boundary; the following measurements quantify its finite-model manifestation and absolute size. In all three families, exact singleton Top-$K$ frequently leaves relative headroom $H_U\ge0.10$, although absolute KL scales and rates vary by family. The global qualifier refers to the enumerated $M=24$ universe.
\begin{table}[H]\centering\small
\caption{Primary exact $M=24,K=4$ audit. Intervals are 95\% prompt-cluster bootstrap intervals. $\Delta\mathrm{KL}=L(S_{\mathrm{rank}})-L(S_U^\star)$.}\label{tab:primary}
\begin{tabular}{lccc}\toprule
Model & $H_U\ge.10$ & Median $H_U$ [95\% CI] & Median $\Delta\mathrm{KL}$ [95\% CI] \\\midrule
SmolLM2 & 55/60 & .926 [.807,.984] & $4.018\cdot10^{-4}$ [$2.491\cdot10^{-4}$,$1.126\cdot10^{-3}$] \\
Gemma-2-2B & 49/60 & .561 [.346,.658] & $1.927\cdot10^{-4}$ [$1.170\cdot10^{-4}$,$3.216\cdot10^{-4}$] \\
Qwen3.5-4B & 60/60 & .954 [.942,.964] & $2.473\cdot10^{-3}$ [$1.544\cdot10^{-3}$,$4.360\cdot10^{-3}$] \\\bottomrule
\end{tabular}
\end{table}
Five outcome-independent candidate universes yield $H_U\ge.10$ rates of 88.1\%, 87.2\%, 86.0\%, 96.4\%, and 99.2\% (Appendix~\ref{app:universes}). In lower-budget exact audits at $K\in\{2,3\}$, rates are 107/120 SmolLM2, 49/120 Gemma, and 118/120 Qwen. The descriptive $K=5$ panel finds 12/12 cells with at least 10\% headroom in each family. A paired SmolLM2 panel compares attention Top-$K$ against exact downstream-singleton Top-$K$: the latter lowers joint-support KL in 38/120 cells, ties in 34, and raises it in 48. This directly illustrates why more accurate marginal scoring does not monotonically improve the final support.

\subsection{Exchange geometry and the reach of the local witness}
\begin{figure}[!htbp]
\centering
\includegraphics[width=.92\linewidth]{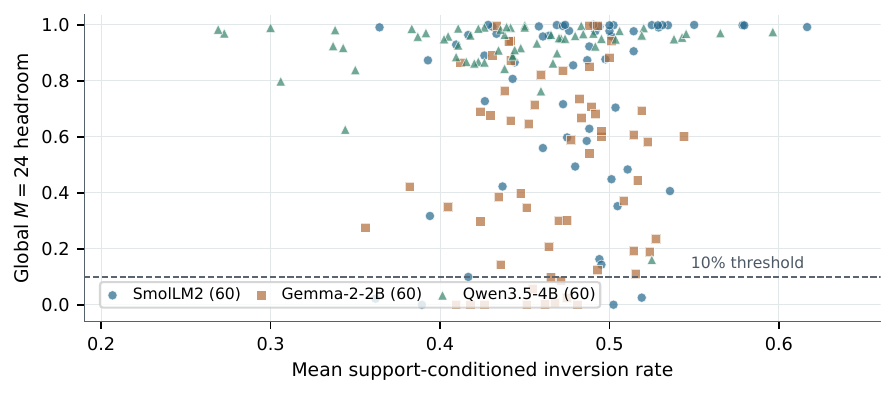}
\caption{\textbf{Support-conditioned ranking instability and joint-support headroom.} Each point is an exact $M=24,K=4$ model/prompt/head cell. The horizontal axis summarizes context-dependent pairwise inversions; the vertical axis is exact candidate-universe headroom. This descriptive plot illustrates the exchange mechanism rather than a population-level causal relation.}
\label{fig:inversions}
\end{figure}

Across the mechanism analysis, 216 cells yield 900 conditioned contexts and 6,512 improving one-swap witnesses. At $K=4$, strict preference reversals occur in 100\% of oracle-positive SmolLM2 cells, 100\% of Gemma cells, and 98.3\% of Qwen cells. The median Kendall $\tau_b$ (tie-adjusted rank correlation) between ranking and conditioned preferences is respectively .025, .062, and .116. Best-one-swap oracle headroom capture medians are 97.16\%, 96.40\%, and 78.10\%, with greedy one-swap search reaching the global oracle in 41.82\%, 69.39\%, and 25.00\% of positive cells, respectively (Appendix~\ref{app:mechanism}). Figure~\ref{fig:inversions} shows the source-level paired measurements across all 180 primary cells.
\paragraph{What the certificate detects.}
\Swap{} fires on 52/55, 48/49, and 57/60 primary supports whose exact candidate-universe headroom satisfies $H_U\ge.10$. These correspond to 157/164 oracle-positive supports pooled descriptively across models; the prompt-cluster 95\% recall intervals are [88.1\%,100\%], [93.2\%,100\%], and [88.3\%,100\%], respectively.
Every firing returns the actual exchanged support and an evaluated loss reduction of at least $10\%$ relative to the incumbent. Among the seven oracle-positive non-firings, six have an improving one-swap below $10\%$, and one Qwen support has global headroom beyond its immediate improving neighborhood. These two mechanisms distinguish \emph{threshold sensitivity} from \emph{one-swap locality} and make the diagnostic outcomes interpretable (Figure~\ref{fig:recall-behavior}; Appendix~\ref{app:swapcert-audit}).

\subsection{Larger-budget multi-layer predictive fidelity}\label{sec:v3-results}
\paragraph{Absolute fidelity at matched active-support budgets.}
Table~\ref{tab:v3-baselines-main} compares four explicit $K$-token selection rules under the same 2,048-token input, dense prefill, and three-layer cached-decode intervention. Attention- and geometric-score supports consistently give smaller mean dense-reference KL than the position-based recent and sink-plus-recent supports in the six tested model/budget settings. For example, at $K=128$, Gemma's mean KL is $6.76\cdot10^{-4}$ under attention selection and $6.31\cdot10^{-4}$ under geometric selection, compared with $4.19\cdot10^{-2}$ for recent and $1.35\cdot10^{-2}$ for sink-plus-recent. These absolute losses ground the paired improvements in the magnitude of the actual predictive distortion.
\begin{table}[H]
\centering
\caption{\textbf{Four fixed-budget baselines at 2K context.} Mean dense-reference next-token KL over 30 paired prompts; lower is better. Every method uses the same active support cardinality $K$ in all heads of the three selected full-attention layers.}
\label{tab:v3-baselines-main}
\small\setlength{\tabcolsep}{4pt}
\begin{tabular}{lrrrrr}\toprule
Model & $K$ & Attention & Geometric & Recent & Sink+recent\\\midrule
SmolLM2 &128& .0011103 & .0009981 & .1794272 & .0425540\\
SmolLM2 &256& .0006785 & .0006863 & .1704914 & .0403637\\
Gemma-2-2B &128& .0006760 & .0006308 & .0419478 & .0134970\\
Gemma-2-2B &256& .0003143 & .0002836 & .0362449 & .0122437\\
Qwen3.5-4B &128& .0011924 & .0011308 & .2587330 & .2472465\\
Qwen3.5-4B &256& .0006270 & .0006127 & .2447003 & .2159831\\\bottomrule
\end{tabular}
\end{table}

\paragraph{Paired KL and gold-answer likelihood.}
Table~\ref{tab:v3-main} adds prompt-paired uncertainty to the geometric-versus-attention comparison and reports \emph{reference-answer negative log likelihood} alongside KL. The NLL is calculated on up to eight answer tokens per prompt, complementing the label-free dense-reference objective with a measured reference-token score. Gemma at $K=128$ has $22/30$ KL wins and $24/30$ NLL wins; its mean KL difference is $-4.52\cdot10^{-5}$ and mean NLL difference is $-0.00411$, with exploratory paired 95\% intervals below zero for both. At $K=256$, Gemma's KL interval also lies below zero. Relative to the corresponding attention-KL means, Gemma's reductions are $6.68\%$ and $9.77\%$ at $K=128$ and $K=256$, respectively. SmolLM2 and Qwen exhibit smaller or more variable differences on their matched prompts. Across models, reporting both metrics reveals how a support's dense-predictive fidelity relates to the likelihood of observed answer tokens.
\begin{table}[H]
\centering
\caption{\textbf{Paired geometric minus attention Top-$K$ differences.} Negative values favor geometric scoring. Each interval is an exploratory prompt-bootstrap 95\% CI ($n=30$; 20,000 resamples). NLL uses up to eight gold answer tokens. The two support budgets reuse each model's prompt panel.}
\label{tab:v3-main}
\scriptsize\setlength{\tabcolsep}{3pt}
\resizebox{\linewidth}{!}{%
\begin{tabular}{lrll}\toprule
Model & $K$ & $\Delta$ dense-reference KL [95\% CI] & $\Delta$ gold-answer NLL [95\% CI] \\\midrule
SmolLM2 &128& $-1.121\cdot10^{-4}$ [$-6.391\cdot10^{-4}$,$+2.739\cdot10^{-4}$] & $-0.00591$ [$-0.01908$,$+0.00860$]\\
SmolLM2 &256& $+7.76\cdot10^{-6}$ [$-1.609\cdot10^{-4}$,$+1.969\cdot10^{-4}$] & $-0.00782$ [$-0.02237$,$+0.00500$]\\
Gemma-2-2B &128& $-4.517\cdot10^{-5}$ [$-8.300\cdot10^{-5}$,$-1.341\cdot10^{-5}$] & $-0.00411$ [$-0.00696$,$-0.00157$]\\
Gemma-2-2B &256& $-3.072\cdot10^{-5}$ [$-5.166\cdot10^{-5}$,$-1.343\cdot10^{-5}$] & $-0.00156$ [$-0.00577$,$+0.00344$]\\
Qwen3.5-4B &128& $-6.155\cdot10^{-5}$ [$-2.167\cdot10^{-4}$,$+6.812\cdot10^{-5}$] & $+0.00569$ [$-0.00023$,$+0.01191$]\\
Qwen3.5-4B &256& $-1.433\cdot10^{-5}$ [$-2.622\cdot10^{-4}$,$+1.811\cdot10^{-4}$] & $-0.00361$ [$-0.01189$,$+0.00458$]\\\bottomrule
\end{tabular}}
\end{table}
\begin{figure}[!htbp]
\centering
\includegraphics[width=.94\linewidth]{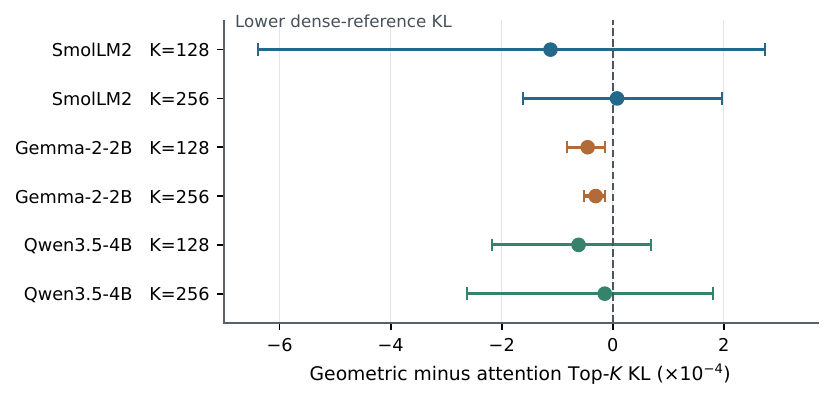}
\caption{\textbf{Paired larger-budget dense-reference KL differences.} Geometric singleton Top-$K$ minus attention Top-$K$ on the same 30 prompts per model; negative favors geometric scoring. Whiskers are exploratory 95\% prompt-bootstrap intervals, unadjusted for multiple comparisons. Both budgets share the same prompt panel.}
\label{fig:v3-kl}
\end{figure}
\FloatBarrier
\paragraph{Constructive exchanges at larger $K$.}
A separate panel evaluates up to eight same-budget exchanges per configuration against equal-count randomized proposals. Positive improving witnesses appear in 39/72 model--budget--prompt configurations, across all three families and both budgets (Appendix~\ref{app:v3-evidence}). The guided-minus-random paired intervals contain zero in each configuration, characterizing the current resolution of the proposal comparison. These observations extend the presence of improving fixed-budget exchanges to the larger-support setting, with the exact candidate-universe oracle reserved for the $M=24$ experiment.

\subsection{Next-token and generated-output behavior}\label{sec:v3-generation}
\begin{figure}[!htbp]
\centering

\includegraphics[width=.98\linewidth]{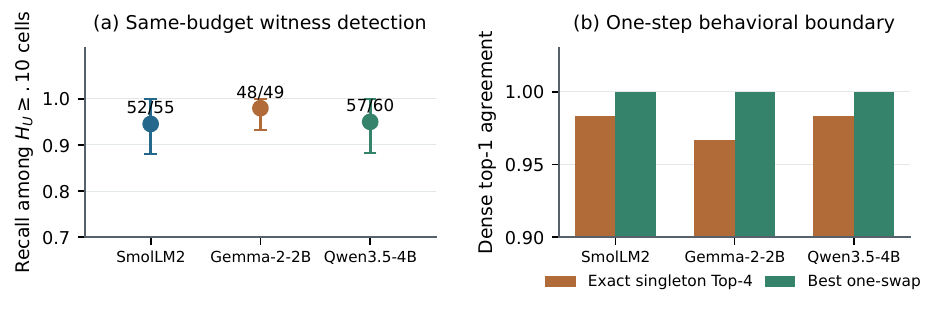}
\caption{\textbf{Constructive diagnosis and observed token decisions at $M=24,K=4$.} Left: \Swap{} recall at $\varepsilon=.10$ against the exact candidate-universe headroom-positive cells (prompt-cluster 95\% intervals). Right: observed agreement with the unmodified dense next-token top-1; all four exact-singleton flips are corrected by the best evaluated same-budget one-swap.}
\label{fig:recall-behavior}
\end{figure}

In the matched $K=4$ next-token behavioral panel, exact-singleton selection changes the dense top-1 in 1/60 SmolLM2, 2/60 Gemma, and 1/60 Qwen cells; in all four, the best same-budget one-swap and the candidate-universe oracle restore the dense decision. The improvement intervals contain zero, consistent with the low frequency of boundary crossings (Figure~\ref{fig:recall-behavior}). Proposition~\ref{prop:margin} clarifies the connection: a small enough dense-reference KL relative to the dense probability margin certifies a stable top-1 prediction.

\paragraph{Certified preservation of the dense top-1.}
We additionally apply Proposition~\ref{prop:margin} to every cell of the $K=4$ next-token behavioral audit. With dense top-token probability $p_1$ and top-two logit difference $\gamma$, the probability margin is computed as $m=p_1(1-e^{-\gamma})$. Each evaluated support satisfying $L(S)<m^2/2$ therefore has a \emph{proved} dense top-1 match. Table~\ref{tab:margin-audit} reports this deterministic test next to observed changes. Across the 180 model--prompt--head cells, the number of mathematically certified matches increases from $152$ with exact-singleton ranking to $159$ after the best same-budget swap; full candidate-universe optimization certifies $161$. All four observed top-1 deviations of the initial ranker are corrected by the best swap. These guarantees are computed at the same frozen prefixes and downstream KL objective as the exact support audit.
\begin{table}[H]
\centering\small
\caption{\textbf{KL--margin certification and observed dense top-1 changes} for the exact $M=24,K=4$ next-token panel. Certified cells satisfy $L(S)<m^2/2$. Each entry is out of 60 model--prompt--head cells.}
\label{tab:margin-audit}
\setlength{\tabcolsep}{4pt}
\begin{tabular}{lrrrr}\toprule
Model & Ranked certified & Best-swap certified & Oracle certified & Ranked flips \\\midrule
SmolLM2 & $56/60$ & $59/60$ & $60/60$ & $1/60$ \\
Gemma-2-2B & $39/60$ & $41/60$ & $41/60$ & $2/60$ \\
Qwen3.5-4B & $57/60$ & $59/60$ & $60/60$ & $1/60$ \\\midrule
Total & $152/180$ & $159/180$ & $161/180$ & $4/180$ \\\bottomrule
\end{tabular}
\end{table}

\FloatBarrier
In the new $K=128$ diagnostic, geometric and attention-ranked supports produce identical 24-token greedy continuations on all 90 paired model--prompt cases (30 per family). They match dense continuations in 30/30 SmolLM2, 28/30 Gemma, and 28/30 Qwen cases. The restricted response budget frequently truncates an answer before a substantive final response: 19/30 SmolLM2 decoded strings are empty and all 30 Qwen strings contain a reasoning tag. The 90 paired continuations therefore characterize decoding stability at this horizon, alongside the gold-token NLL and KL measurements; Appendix~\ref{app:v3-evidence} records the task-output details.

\section{Implications and Discussion}\label{sec:discussion}
\paragraph{A general principle for sparse attention.}
The unit of optimization matters. An individual relevance score asks how much a candidate influences the dense computation under a \emph{singleton} intervention. Joint support selection asks which $K$ candidates produce the smallest downstream deviation after \emph{collective} renormalization. The finite counterexample, unbounded-ratio family, and all-budget preference reversals establish a sharp mathematical separation between the two. The gap persists even when the singleton score is evaluated exactly under the eventual set-level objective. This provides a clear reason to evaluate both the \emph{marginal ranker} and the \emph{jointly selected support}.

\paragraph{When a support-quality gap becomes useful.}
The exhaustive oracle makes relative headroom measurable, while absolute downstream KL quantifies its predictive scale. A high relative gap can arise from small absolute KL values, so both are reported in the main experiments. \Swap{} adds a local, inspectable certificate: the proposed outgoing and incoming positions, their evaluated support losses, and the exact size-$K$ replacement. Its $K(M-K)$ candidate evaluations provide a controlled audit cost; the $\binom{M}{K}$ oracle provides independent calibration. The standard neighborhood enumeration is deliberately transparent. Its scientific role is to expose support-level information that a singleton ranking alone cannot certify.

\paragraph{From KL to decisions and generated answers.}
The expected-log-score identity connects the loss directly to predictive distributions at open-ended prefixes. The KL--margin test certifies dense top-1 preservation in 152/180 exact-singleton supports, increasing to 159/180 after the best one-swap and 161/180 under the candidate-universe optimum. In the four observed rank-induced top-token changes, the best one-swap restores the dense decision. The trajectory chain rule gives a formal route from conditional KL along dense continuations to sequence-event stability. At the larger $K\in\{128,256\}$ budgets, paired gold-answer NLL offers a reference-token metric beside the dense-reference KL. The short greedy test shows that probability changes and discrete outputs need not move together: attention and geometric rankings produce identical short continuations in the matched 90 cases. This combination defines a principled assessment hierarchy: support loss, token decisions, sequence-distribution fidelity, and actual answer quality.

\paragraph{A broader open problem: budgeted, downstream-aware support policies.}
In deployed inference, supports vary by layer, head, prefix, and generation time. A natural research target is a support policy $\pi$ selecting $S_{\ell,h,t}$ at layer $\ell$, head $h$, and decoding step $t$, under a resident-memory budget $B$ and runtime budget $\tau$:
\begin{equation}\label{eq:research-frontier}
\min_{\pi}\;\sum_{t=1}^{T}\E_{Y_{<t}\sim P_D}\!\left[\KL\bigl(p_D(\cdot\mid Y_{<t})\Vert p_\pi(\cdot\mid Y_{<t})\bigr)\right],
\quad \text{s.t.}\quad C_{\mathrm{bytes}}(\pi)\le B,\; C_{\mathrm{time}}(\pi)\le\tau.
\end{equation}
Here the conditional predictor includes all interacting layers, $C_{\mathrm{bytes}}(\pi)$ measures actual resident KV bytes, and $C_{\mathrm{time}}(\pi)$ includes scoring, support selection, and inference time. This formulation separates the quality of a support policy from the cost of constructing and deploying it. It connects to submodular selection, value-aware scoring, adaptive budget allocation, and systems-level KV management, but imposes the downstream predictive objective and sequence coupling explicitly.

The structural results motivate three concrete challenges: finding economical surrogates for support-conditioned loss; deriving useful bounds or pruning certificates that improve on direct neighborhood enumeration; and validating such policies with physically enforced cache budgets, inference cost, and full-answer evaluation. The present exact and cached-decode audits make these challenges quantitatively testable rather than merely conceptual.

\section{Conclusion}
Exact downstream singleton importance is an informative measurement of individual relevance, yet it does not generally approximate the best jointly renormalized fixed-budget support: the multiplicative gap can be unbounded. We prove this boundary, derive the exchange mechanism and all-budget preference reversals, and measure its finite-model consequences through exhaustive downstream-KL audits. \Swap{} turns remaining candidate-universe headroom into a checkable same-budget witness, while KL--margin certification and larger-budget cached decoding link support quality to predictive behavior. The resulting foundation places \emph{joint downstream support quality} alongside ranking accuracy as an essential object of study for sparse LLM inference. It opens a precise agenda for support-aware algorithms and end-to-end evaluation under real memory, compute, and generation constraints.

\section*{Reproducibility and Use of Computational Tools}
All reported original experiments use the fixed protocols and candidate universes detailed in the appendices. The accompanying numerical artifacts include the exact counterexample verifier, raw model/prompt/head audit records, larger-budget outputs, figure-source data, and the KL--margin audit. Automated tools, including generative AI systems, assisted literature discovery, editing, source-code inspection, and proof cross-validations. The author is responsible for the mathematical and empirical claims.

\appendix
\section{Exact Constructions and Proof Details}\label{app:counterexample}
\subsection{Every singleton-deletion and size-two support}
Let $a=(0.2,0.1,0.2,0.5)$, $v=(4,-2,1,2.7)$ and $h_D=\sum_j a_jv_j=2.15$. Deleting candidate $j$ and renormalizing gives $h_{-j}=(h_D-a_jv_j)/(1-a_j)$. With the logistic sigmoid $\sigma(t)=1/(1+e^{-t})$, define $q(x)=\sigma(1.8x)$ and
\[
D(x)=q(h_D)\log\frac{q(h_D)}{q(x)}+[1-q(h_D)]\log\frac{1-q(h_D)}{1-q(x)}.
\]
The exact singleton scores $d_j=D(h_{-j})$ are:
\begin{center}\begin{tabular}{ccc}\toprule
$j$ & $h_{-j}$ & $d_j$ \\\midrule
1&1.687500&0.00918668148\\
2&2.611111&0.00536906759\\
3&2.437500&0.00228497479\\
4&1.600000&0.01374411597\\\bottomrule
\end{tabular}\end{center}
The jointly renormalized size-two supports yield:
\begin{center}\begin{tabular}{ccc}\toprule
$S$&$h_S$&$L(S)$\\\midrule
$\{1,2\}$&2.000000&0.000796590380\\
$\{1,3\}$&2.500000&0.003276210803\\
$\{1,4\}$&3.071429&0.017208222670\\
$\{2,3\}$&0.000000&0.593430803601\\
$\{2,4\}$&1.916667&0.002026829576\\
$\{3,4\}$&2.214286&0.000129168300\\\bottomrule
\end{tabular}\end{center}
Ranking selects $\{1,4\}$; exact minimization selects $\{3,4\}$, with ratio $0.017208222670/0.000129168300\simeq133.2$.
\subsection{Unbounded multiplicative suboptimality: exact construction}\label{app:unbounded-proof}
The following perturbation family establishes Proposition~\ref{prop:unbounded} analytically. Use fixed weights
\(a=(1/5,1/10,1/5,1/2)\), values
\(v(\delta)=(4,-2,1/4+\delta,27/10)\), and binary downstream logits
\(z(h)=(\lambda h,0)\) with \(\lambda=9/5\).
Then
\begin{equation}\label{eq:unbound-outputs}
h_D(\delta)=2+\delta/5,\quad
h_{\{1,4\}}(\delta)=43/14,\quad
h_{\{3,4\}}(\delta)=2+2\delta/7,
\end{equation}
so \(h_{\{3,4\}}-h_D=3\delta/35\). When \(\delta=0\), \(h_{\{3,4\}}=h_D=2\), giving support loss zero, whereas \(h_{\{1,4\}}=43/14\ne2\), giving positive loss
\(L(\{1,4\})=0.02830090151079275\).

The singleton-deletion outputs at \(\delta=0\), in the order \(j=4,1,2,3\), are
\[
h_{-4}=13/10,\qquad h_{-1}=3/2,\qquad h_{-2}=22/9,\qquad h_{-3}=39/16.
\]
Write \(A(t)=\log(1+e^t)\) and \(t_D=\lambda h_D=18/5\). The exact Bernoulli KL of a sigmoid-logit shift from \(t_D\) to \(t\) is the Bregman expression
\[
D(t_D,t)=A(t)-A(t_D)-A'(t_D)(t-t_D).
\]
Since \(A''(t)=\sigma(t)(1-\sigma(t))\) strictly decreases for positive \(t\), a negative shift of magnitude \(u>0\) produces greater KL than a positive shift of the same magnitude from \(t_D>0\), as long as the shifted logits remain positive. The negative shifts for deletions 4 and 1 have magnitudes \(1.26\) and \(0.9\); the positive shifts for deletions 2 and 3 have magnitudes \(0.8\) and \(0.7875\). Monotonicity of the Bregman expression along each side therefore gives the strict exact singleton ordering
\begin{equation}\label{eq:unbound-singleton-order}
d_4>d_1>d_2>d_3.
\end{equation}
For reference, the KL values in the same order are
\[
(d_4,d_1,d_2,d_3)\approx
(0.03149678,\ 0.01414917,\ 0.00652309,\ 0.00634317).
\]

The ordering in Eq.~\eqref{eq:unbound-singleton-order} persists for all sufficiently small \(\delta>0\) by continuity. For such \(\delta\), the ranked support remains \(\{1,4\}\), and its KL converges to \(0.02830090151079275>0\) as \(\delta\downarrow0\). Meanwhile
\[
0<\min_{|S|=2}L(S)\le L(\{3,4\})\longrightarrow0.
\]
To justify strict positivity of the left side, observe that at \(\delta=0\) the only size-two supports reproducing \(h_D\) are \(\{1,2\}\) and \(\{3,4\}\). For \(\delta>0\), their output differences from \(h_D\) are \(-\delta/5\) and \(3\delta/35\), both nonzero; the other four differences remain nonzero in a neighborhood of zero. The downstream sigmoid is strictly monotone, so zero KL is equivalent to equal head outputs. The optimal positive loss can therefore be arbitrarily small while ranked loss remains bounded below. This proves the unbounded ratio with strictly positive numerator and denominator.

\begin{table}[htbp]\centering\small
\caption{Illustration of the analytically established perturbation family. The support ranking is \(\{1,4\}\) and the best pair is \(\{3,4\}\) at each listed positive \(\delta\). The unbounded conclusion follows from the proof, not from finite numerical search.}\label{tab:unbound-numerics}
\begin{tabular}{rrrr}\toprule
$\delta$ & $L(\{1,4\})$ & $L(\{3,4\})$ & Loss ratio \\\midrule
$10^{-1}$ & $0.02655013167$ & $2.963552881\cdot10^{-6}$ & $8{,}958.9$ \\
$10^{-2}$ & $0.02812162715$ & $3.069408823\cdot10^{-8}$ & $916{,}190.3$ \\
$10^{-3}$ & $0.02828293148$ & $3.080189875\cdot10^{-10}$ & $91{,}822{,}039.0$ \\\bottomrule
\end{tabular}
\end{table}

\subsection{Exchange and reversal calculations}
For $S'=S\setminus\{i\}\cup\{j\}$, substituting $\sum_{r\in S'}a_r v_r=A_Sh_S-a_iv_i+a_jv_j$ and $A_{S'}=A_S-a_i+a_j$ proves Eqs.~\eqref{eq:exchange1}--\eqref{eq:exchange2} directly. In Proposition~\ref{prop:reversal}, the dense output is zero. With symmetric binary logits $(z,0)$, loss is $\log\cosh(z/2)$, increasing strictly with $|z|$. The two contexts $C_-$ and $C_+$ formed from shared zeros plus $-2$ or $+2$ yield completed support means $(-1/K,-3/K)$ and $(3/K,1/K)$ for candidates $+1$ and $-1$. The preference reverses in the two contexts.
\subsection{One-sided soundness}
For any $S\in\cF_K(U)$, $\cN_1(S)\subset\cF_K(U)$, hence $\min_{T\in\cF_K(U)}L(T)\le\min_{T\in\cN_1(S)}L(T)$. Dividing by $L(S)>0$ gives $H_U(S)\ge W_U(S)$. At $L(S)=0$, the quantities are defined to be zero. This proves Theorem~\ref{thm:soundness}.

\section{Autoregressive Predictive-Fidelity Bridge}\label{app:sequence-bridge}
Consider a fixed initial prompt and two autoregressive conditional predictors whose conditional next-token distributions are defined on every continuation prefix. Denote the dense predictor by \(p_D(y_t\mid y_{<t})\) and the support-policy predictor by \(p_\pi(y_t\mid y_{<t})\). Their induced finite-length laws are
\[
P_D^{(T)}(y_{1:T})=\prod_{t=1}^T p_D(y_t\mid y_{<t}),\qquad
P_\pi^{(T)}(y_{1:T})=\prod_{t=1}^T p_\pi(y_t\mid y_{<t}).
\]
Assume finite vocabulary and strictly positive softmax probabilities. Expanding the log ratio into a sum of conditional log ratios and taking expectation with respect to \(P_D^{(T)}\) gives the chain-rule identity in Eq.~\eqref{eq:trajectory-kl}. In particular, if the dense-prefix expected sum of conditional KL divergences is at most \(\epsilon\), Pinsker's inequality yields \(|P_D^{(T)}(A)-P_\pi^{(T)}(A)|\le\sqrt{\epsilon/2}\) for every event \(A\) on length-$T$ continuations. The guarantee concerns the probability of an event under the two \emph{generative distributions}; a benchmark prediction is evaluated by the actual sampled or decoded response and its task-specific scoring rule. The conditional KL in this identity is evaluated on prefixes sampled from the dense sequence law, so independent one-step prefix probes serve as empirical building blocks rather than a substitute for the full expected sum.

\section{Experimental Definitions and Statistical Units}\label{app:definitions}
Every primary exhaustive statement refers to $M=24$, $K=4$ within a frozen candidate universe. Each model has 30 prespecified LongBench-E prompts, 10 from each of 2WikiMultihopQA, HotpotQA, and PassageRetrieval-en. The first and last full-softmax heads are examined at the deepest eligible softmax layer: zero-indexed 23 for SmolLM2, 25 for Gemma and 31 for Qwen. Each prompt contributes two correlated head measurements; prompt-cluster bootstrap resamples prompts with their heads together. The 10\% threshold and primary evaluation design were fixed before the primary outcomes.

\section{Candidate-Universe Sensitivity}\label{app:universes}
Let $W_o$ denote the attention head\'s output-projection matrix. Candidate universes are formed independently of downstream support-loss outcomes: U1 retains the top 16 attention positions and 8 remaining positions with the largest $a_j\|W_o(v_j-h_D)\|_2$; U2 uses the top 24 attention positions; U3 uses top 12 attention plus 12 uniformly sampled remaining valid positions (two seeds); U4 samples 24 valid positions uniformly (two seeds); U5 samples 6 from each of four contiguous attention-rank quartiles.
\begin{table}[htbp]\centering\small
\caption{Exact $M=24$ candidate-universe sensitivity, across three models.}\label{tab:universe}
\begin{tabular}{lrrcc}\toprule
Construction&Cells&Rate $H_U\ge.10$&95\% cluster CI&Median $H_U$\\\midrule
U1 hybrid&360&.8806&[.8250,.9306]&.8329\\
U2 attention&360&.8722&[.8167,.9222]&.8275\\
U3 mixed, two seeds&720&.8597&[.8028,.9139]&.8272\\
U4 random, two seeds&720&.9639&[.9389,.9847]&.8851\\
U5 stratified&360&.9917&[.9750,1.0000]&.8948\\\bottomrule
\end{tabular}\end{table}

\section{Exact Lower-Budget Replication}\label{app:lowerK}
\begin{table}[htbp]\centering\small
\caption{Exact $M=24$, $K\in\{2,3\}$ cross-model audit (120 cells/model).}\label{tab:lowerK}
\begin{tabular}{lcccc}\toprule
Model&$H_U\ge.10$&95\% cluster CI&Median $H_U$&Median $\Delta$KL\\\midrule
SmolLM2&107/120&[.825,.950]&.83058&$3.7795\cdot10^{-4}$\\
Gemma&49/120&[.325,.4917]&.004078&$1.6688\cdot10^{-6}$\\
Qwen&118/120&[.950,1.000]&.92994&$3.0970\cdot10^{-3}$\\\bottomrule
\end{tabular}\end{table}
The Gemma lower-$K$ behavior differs markedly from its $K=4$ results, underscoring model--budget interactions.

\section{Primary \texorpdfstring{$K=4$}{K=4} Statistical Audit}\label{app:primary-statistics}
Cluster-bootstrap median global $M=24$ headroom estimates are SmolLM2 .92591 [.80678,.98388], Gemma .56094 [.34624,.65773], and Qwen .95385 [.94152,.96414]. Median absolute-KL improvement estimates are respectively $4.0183\cdot10^{-4}$ [$2.4910\cdot10^{-4}$,$1.12597\cdot10^{-3}$], $1.9266\cdot10^{-4}$ [$1.1702\cdot10^{-4}$,$3.2155\cdot10^{-4}$], and $2.4732\cdot10^{-3}$ [$1.5441\cdot10^{-3}$,$4.3597\cdot10^{-3}$]. Independent recomputation of the primary K=4 support losses has maximum absolute discrepancy zero.

\section{Support-Conditioning Mechanism Audit}\label{app:mechanism}
Across 216 cells, the mechanism audit evaluates 900 conditioned contexts and records 6,512 improving one-exchange witnesses.
\begin{table}[htbp]\centering\small
\caption{Support-conditioned mechanism at $K=4$; reversal and greedy figures refer to cells with $H_U\ge.10$.}\label{tab:mechanism}
\begin{tabular}{lccccc}\toprule
Model&Reversal&Median $\tau_b$&Median inversion&Best-swap capture&Greedy oracle\\\midrule
SmolLM2&100\%&.0250&48.75\%&97.16\%&41.82\%\\
Gemma&100\%&.0619&46.90\%&96.40\%&69.39\%\\
Qwen&98.33\%&.1155&44.23\%&78.10\%&25.00\%\\\bottomrule
\end{tabular}\end{table}
High one-swap capture can coexist with incomplete greedy convergence because a single strong exchange and a path to the global optimum are different properties.

\section{SwapCert Positive-Witness Accounting}\label{app:swapcert-audit}
\begin{table}[htbp]\centering\small
\caption{All $K=4$, $\varepsilon=.10$ positive-certificate outcomes.}\label{tab:swapcert-recall}
\begin{tabular}{lcccc}\toprule
Model&Oracle positives&Hits&Misses&Recall [95\% CI]\\\midrule
SmolLM2&55&52&3&94.55\% [88.14\%,100\%]\\
Gemma&49&48&1&97.96\% [93.18\%,100\%]\\
Qwen&60&57&3&95.00\% [88.33\%,100\%]\\\bottomrule
\end{tabular}\end{table}
The 157/164 pooled recall is descriptive. The seven misses consist of six cases with positive one-swap gain below threshold and one Qwen cell whose best one-swap is not improving despite more than 10\% global candidate-universe headroom. All issued certificates are sound by Theorem~\ref{thm:soundness}.

\section{Next-Token Behavioral Audit and Numerical Consistency}\label{app:behavior}
The original one-step audit contains 180 cells. Exact-singleton Top-4 changes dense top-1 in one SmolLM2, two Gemma, and one Qwen cells. Each is repaired by the best one-swap and the exact candidate-universe oracle. Prompt-cluster agreement-gain intervals are [0,5\%], [0,8.33\%], and [0,5\%]. In the primary bfloat16 (BF16) implementation, batch-shape dependence affected four Gemma behavioral cells (maximum KL discrepancy $9.1527\cdot10^{-5}$ relative to the primary calculation's $2\cdot10^{-7}$ tolerance). Recomputing behavioral quantities at batch size one and independently reproducing primary exhaustive scores preserved the original $K=4$ values exactly.

\section{Descriptive \texorpdfstring{$K=5$}{K=5} Stress Test}\label{app:k5}
The fixed descriptive extension includes 12 cells/model. All 12/12 cells per model exhibit candidate-universe headroom of at least 10\%. Median $H_U$ is .9751 SmolLM2, .6179 Gemma, and .9364 Qwen. These observations complement the primary $K=4$ analysis.

\section{Larger-Context Swap Screens and Marginal Ranking}\label{app:larger-screens}
A separate exact SmolLM2 panel uses an outcome-independent hybrid $M=24$ universe (top 16 attention positions, plus eight geometric-leverage positions), 30 prompts, two heads, and $K\in\{2,3\}$, comparing attention, head-output singleton magnitude, and exact downstream singleton KL. Another approximately 2K-token swap screen spans three models, three layer-depth settings, two sparsity settings, 1,512 heads, and 48,384 prespecified exchanges; an approximately 4K-token last-layer replication spans 504 heads and 16,128 exchanges. Each head is tested against 32 same-budget exchanges. These report observed improving swaps rather than candidate-universe global optima. In the paired SmolLM2 exact panel, exact downstream-singleton Top-$K$ has lower support KL than attention Top-$K$ in 38 cells, ties in 34 and higher KL in 48; headroom $H_U\ge.10$ occurs in 108/120 exact-singleton and 104/120 attention-ranked supports.

\section{Reproducibility and Protocol Integrity}\label{app:reproducibility}
The supplementary source package includes numerical verifiers for the exact counterexample and reversal construction, machine-readable primary enumeration and sensitivity evidence, the $K=4$ witness audit, and the secondary cached-decode scripts and raw prompt-level results. The primary results use the frozen prompt/model/head/budget settings stated above. Separate output files and analysis rules distinguish exact enumeration, descriptive supporting evaluations, and larger-budget controls; protocol-dependent model precision and native-attention backend details are recorded with each measurement.
\section{Controlled multi-layer, larger-budget cached-decode audits}
\label{app:v3-evidence}

\paragraph{Purpose and experimental unit.}
The primary paper exhaustively evaluates a single full-softmax head over $M=24$ candidates at small $K$. This appendix describes an additional fixed-budget comparison at 2,048-token prompt length. The common paired unit is a prompt, not an individual decoded token, attention head, layer, support budget, or swap. The 30 LongBench-E prompts comprise 10 examples each from 2WikiMultihopQA, HotpotQA, and PassageRetrieval-en. Prompt hashes are matched across methods and budgets. The model revisions, experimental scripts, and raw JSON Lines (JSONL) evidence are distributed separately with the supplementary artifact.

\paragraph{Intervention and ranking rules.}
A standard dense prefill produces cached states. Each subsequent single-query decode invocation is evaluated with the last three eligible full-softmax attention layers sparsified at all heads. Earlier layers and non-softmax blocks remain as in the frozen model. Support selection is recomputed per active head and decode step using the same budget $K=128$ or $256$. In addition to the unmodified dense path, we compare attention-weight Top-$K$, geometric singleton Top-$K$ $u_j=a_j\|v_j-h_D\|_2/(1-a_j)$, recent-token selection, and sink-plus-recent selection. The geometric score measures an exact singleton \emph{head-output} displacement norm; the original exact-enumeration study uses the downstream singleton KL $d_j$ of Eq.~(4). All surviving softmax weights are renormalized. The intervention modifies per-head active attention supports over cached values; cache residency is held fixed. Effective active budget and actual resident bytes are thus separate quantities in this protocol.

\paragraph{Controls, precision, and uncertainty.}
The controlled decode panel computes dense-reference next-token KL and observed gold-answer NLL for up to eight answer tokens, with 30 prompts per model/budget setting. Gemma uses half-precision floating point (FP16) after a dense comparison between eager attention and scaled dot-product attention (SDPA); SmolLM2 and Qwen core results use BF16. Independent Qwen native-dtype/FP16 generation is kept separate. Head-output reconstruction is checked by comparing the reconstructed full-attention head output with the model's native output; backend parity is checked at the prompt level in the strict runs. Exploratory 95\% confidence intervals below use 20,000 prompt-level paired percentile-bootstrap replicates with fixed seed 2027. The intervals are descriptive, unadjusted for the six tested configurations; both budgets share the same prompt panel.

\begin{table}[t]
\centering
\caption{Controlled 2K core results. Both $\Delta$ columns are geometric-singleton Top-$K$ minus attention Top-$K$; negative values favor geometric ranking. Intervals are exploratory prompt-bootstrap 95\% CIs.}
\label{tab:v3-full-core}
\scriptsize
\setlength{\tabcolsep}{3pt}
\resizebox{\linewidth}{!}{%
\begin{tabular}{llrlll}
\toprule
Model & Precision & $K$ & $\Delta$ dense-reference KL [95\% CI] & $\Delta$ gold NLL [95\% CI] & KL wins \\
\midrule
SmolLM2 & BF16 &128 & $-0.00011213$ [$-0.00063914,0.00027391$] & $-0.0059069$ [$-0.019080,0.008597$] & 17/30\\
SmolLM2 & BF16 &256 & $+0.00000776$ [$-0.00016088,0.00019693$] & $-0.0078218$ [$-0.022374,0.005003$] & 17/30\\
Gemma-2-2B & FP16 &128 & $-0.00004517$ [$-0.00008300,-0.00001341$] & $-0.0041058$ [$-0.006957,-0.001570$] & 22/30\\
Gemma-2-2B & FP16 &256 & $-0.00003072$ [$-0.00005166,-0.00001343$] & $-0.0015625$ [$-0.005772,0.003444$] & 24/30\\
Qwen3.5-4B & BF16 &128 & $-0.00006155$ [$-0.00021672,0.00006812$] & $+0.0056863$ [$-0.000230,0.011912$] & 18/30\\
Qwen3.5-4B & BF16 &256 & $-0.00001433$ [$-0.00026221,0.00018111$] & $-0.0036081$ [$-0.011886,0.004577$] & 16/30\\
\bottomrule
\end{tabular}}
\end{table}

\paragraph{Additional selection baselines.}
The full four-method absolute-KL comparison appears in Table~\ref{tab:v3-baselines-main} of the main text. Recent-only and sink-plus-recent selectors use the same head/layer intervention, prompt panel, and active-support budget as attention and geometric selection. The full core paired KL/NLL table above additionally records numerical precision and KL wins per configuration.

\paragraph{Exchange-witness study.}
For each of the six model/budget settings, we evaluate a separate set of 12 prompts and at most eight prespecified exchange candidates per configuration, including equal-budget randomized controls. Strictly improving one-swap witnesses occur in 7/12, 7/12, 8/12, 7/12, 5/12, and 5/12 prompt configurations (SmolLM2 $K=128,256$; Gemma $K=128,256$; Qwen $K=128,256$), respectively: 39/72 total tested configurations. The 95\% intervals for best-guided-minus-best-randomized KL include zero in each configuration. Together these measurements document both the availability of positive same-budget exchanges and the sampling uncertainty in proposal selection. Gemma witness experiments use BF16 and are analyzed separately from Gemma FP16 core experiments.

\paragraph{24-token greedy-generation diagnostic.}
At $K=128$, 30 prompts per model are decoded for up to 24 greedy tokens. Dense, attention Top-$K$, and geometric Top-$K$ are paired within each prompt. The geometric and attention selectors yield the same decoded continuation in 30/30 SmolLM2, 30/30 Gemma, and 30/30 Qwen prompts (90/90 total). This is an exact observed equality on the frozen 24-token panel, directly indicating that the paired KL/NLL changes did not alter these short greedy output strings. Reported answer exact match is zero for every selector on this restricted panel; mean geometric/attention question-answering token-overlap F1 (QA-F1, the harmonic mean of token precision and recall) is 0.0425, 0.1154, and 0.0656, respectively. Of the short decoded outputs, 19/30 SmolLM2 outputs are empty strings and 30/30 Qwen outputs contain a \texttt{<think>} segment. These protocol features affect the interpretation of short-answer scores; the unchanged continuation finding directly describes only this 24-token diagnostic. Qwen generation uses FP16 with the validated native-dtype reconstruction path.

\paragraph{Relationship to the exact theory and original audits.}
Theorem~1 and the objective-matched counterexample address the mathematical gap between exact downstream singleton ranking and the combinatorial optimum of a fixed universe. The 2K extension measures a related output-geometric singleton rule at larger budgets during cached decoding; the exact candidate-universe oracle and one-sided certificates are evaluated in the $M=24$ study. The two protocols therefore provide complementary observations at their respective experimental scales. The appendix reports precision-specific numerical controls and the full outcomes, including near-ties, to facilitate independent interpretation.

\section{KL--Margin Decision Certification on the Exact Behavioral Panel}\label{app:margin-certification}
For each of the 180 $K=4$ next-token model--prompt--head cells, let $p_1$ be the dense top-token probability and let $\gamma$ be the recorded difference between the largest and second-largest dense logits. The dense probability margin is $m=p_1(1-e^{-\gamma})$, since $p_2=p_1e^{-\gamma}$. Applying Proposition~\ref{prop:margin} to each saved support gives the deterministic criterion $\KL(p_D\Vert p_S)<m^2/2$. Within the SmolLM2/Gemma/Qwen panels of 60 cells each, this criterion holds for $56/39/57$ exact-singleton ranked supports, $59/41/59$ best one-swap supports, and $60/41/60$ exact candidate-universe oracle supports. The corresponding pooled counts are $152/180$, $159/180$, and $161/180$. The ranked supports change the dense top-1 in $1/2/1$ cells by model; best swaps and oracles preserve it throughout the panel. The supplementary program \texttt{margin\_certificate\_audit.py} recomputes each margin, tests all three strict inequalities, confirms agreement for every certified cell, and writes the complete 180-row derived table. The counts refer to fixed recorded interventions and do not require statistical estimation of a model population.

\end{document}